\documentclass[runningheads]{llncs}
\usepackage{graphicx}
\usepackage{tikz}
\usepackage{comment}
\usepackage{amsmath,amssymb} % define this before the line numbering.
\usepackage{color}
\usepackage{multirow}
\usepackage{wrapfig}
\usepackage{booktabs}
\usepackage[nice]{nicefrac}
\usepackage{floatrow}
 \usepackage{nohyperref}

\newfloatcommand{capbtabbox}{table}[][\FBwidth]

\usepackage{makecell}
\usepackage{blindtext}
\usepackage{nicefrac}

\usepackage[accsupp]{axessibility}  % Improves PDF readability for those with disabilities.

\newtheorem{prop}{Proposition}
\newtheorem{duplicate}{Proposition}

\begin{document}
% \linenumbers
\pagestyle{headings}
\mainmatter
\def\ECCVSubNumber{2041}  % Insert your submission number here

\title{Improving Covariance Conditioning of the SVD Meta-layer by Orthogonality} % Replace with your title

\begin{comment}
\titlerunning{ECCV-22 submission ID \ECCVSubNumber} 
\authorrunning{ECCV-22 submission ID \ECCVSubNumber} 
\author{Anonymous ECCV submission}
\institute{Paper ID \ECCVSubNumber}
\end{comment}
%******************

%\begin{comment}
\titlerunning{Improving Conditioning by Orthogonality}
\author{Yue Song\orcidID{0000-0003-1573-5643} \and
Nicu Sebe \and
Wei Wang}
%\orcidID{2222--3333-4444-5555}
\authorrunning{Y. Song et al.}
\institute{DISI, University of Trento, Trento 38123, Italy\\
\email{yue.song@unitn.it}\\
\url{https://github.com/KingJamesSong/OrthoImproveCond}
}

%\end{comment}
%******************
\maketitle

\begin{abstract}
Inserting an SVD meta-layer into neural networks is prone to make the covariance ill-conditioned, which could harm the model in the training stability and generalization abilities. In this paper, we systematically study how to improve the covariance conditioning by enforcing orthogonality to the Pre-SVD layer. Existing orthogonal treatments on the weights are first investigated. However, these techniques can improve the conditioning but would hurt the performance. To avoid such a side effect, we propose the Nearest Orthogonal Gradient (NOG) and Optimal Learning Rate (OLR).  The effectiveness of our methods is validated in two applications: decorrelated Batch Normalization (BN) and Global Covariance Pooling (GCP). Extensive experiments on visual recognition demonstrate that our methods can simultaneously improve the covariance conditioning and generalization. Moreover, the combinations with orthogonal weight can further boost the performances. 

%Under our simplification, manipulating only one layer previous to the SVD layer can attain much better covariance conditioning and bring obvious generalization improvements. We 

%Applying different constraint to the weight, gradient, and even learning rate of that layer 

\end{abstract}

\def\ma{{\mathbf{a}}}
\def\ml{{\mathbf{l}}}
\def\mw{{\mathbf{w}}}
\def\mi{{\mathbf{i}}}
\def\mA{{\mathbf{A}}}
\def\mB{{\mathbf{B}}}
\def\mC{{\mathbf{C}}}
\def\mD{{\mathbf{D}}}
\def\mE{{\mathbf{E}}}
\def\mF{{\mathbf{F}}}
\def\mG{{\mathbf{G}}}
\def\mH{{\mathbf{H}}}
\def\mI{{\mathbf{I}}}
\def\mJ{{\mathbf{J}}}
\def\mK{{\mathbf{K}}}
\def\mL{{\mathbf{L}}}
\def\mM{{\mathbf{M}}}
\def\mN{{\mathbf{N}}}
\def\mO{{\mathbf{O}}}
\def\mP{{\mathbf{P}}}
\def\mQ{{\mathbf{Q}}}
\def\mR{{\mathbf{R}}}
\def\mS{{\mathbf{S}}}
\def\mT{{\mathbf{T}}}
\def\mU{{\mathbf{U}}}
\def\mV{{\mathbf{V}}}
\def\mW{{\mathbf{W}}}
\def\mX{{\mathbf{X}}}
\def\mY{{\mathbf{Y}}}
\def\mZ{{\mathbf{Z}}}
\def\mBeta{{\mathbf{\beta}}}
\def\mPhi{{\mathbf{\Phi}}}
\def\mLambda{{\mathbf{\Lambda}}}
\def\mSigma{{\mathbf{\Sigma}}}

\section{Introduction}

% Identify the phenomenon of ill-conditioning
The Singular Value Decomposition (SVD) can factorize a matrix into orthogonal eigenbases and non-negative singular values, serving as an essential step for many matrix operations. Recently in computer vision and deep learning, many approaches integrated the SVD as a meta-layer in the neural networks to perform some differentiable spectral transformations, such as the matrix square root and inverse square root. The applications arise in a wide range of methods, including Global Covariance Pooling (GCP)~\cite{li2017second,song2021approximate,gao2021temporal}, decorrelated Batch Normalization (BN)~\cite{huang2018decorrelated,huang2021group,song2022fast}, Whitening an Coloring Transform (WCT) for universal style transfer~\cite{li2017universal,chiu2019understanding,wang2020diversified}, and Perspective-n-Point (PnP) problems~\cite{brachmann2017dsac,campbell2020solving,dang2020eigendecomposition}.

For the input feature map $\mX$ passed to the SVD meta-layer, one often first computes the covariance of the feature as $\mX\mX^{T}$. This can ensure that the covariance matrix is both symmetric and positive semi-definite, which does not involve any negative eigenvalues and leads to the identical left and right eigenvector matrices. However, it is observed that inserting the SVD layer into deep models would typically make the covariance very ill-conditioned~\cite{song2021approximate}, 
resulting in deleterious consequences on the stability and optimization of the training process. For a given covariance $\mA$, its conditioning is measured by the condition number:
\begin{equation}
    \kappa(\mA) = \sigma_{max}(\mA) \sigma_{min}^{-1}(\mA) %\nicefrac{\sigma_{max}(\mA)}{}
\end{equation}
where $\sigma(\cdot)$ denotes the eigenvalue of the matrix. Mathematically speaking, the condition number measures how sensitive the SVD is to the errors of the input. Matrices with low condition numbers are considered \textbf{well-conditioned}, while matrices with high condition numbers are said to be \textbf{ill-conditioned}. Specific to neural networks, the ill-conditioned covariance matrices are harmful to the training process in several aspects, which we will analyze in detail later.

\begin{figure}[htbp]
\vspace{-0.4cm}
    \centering
    \includegraphics[width=0.8\linewidth]{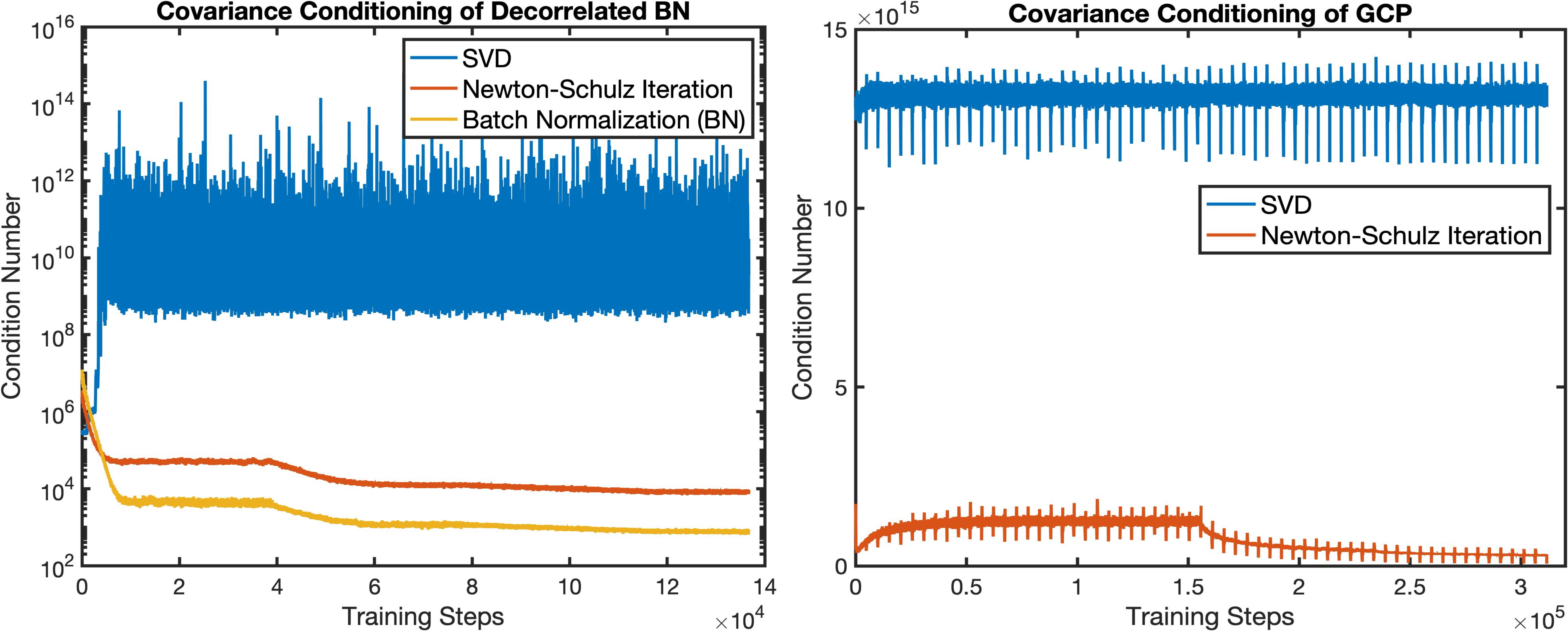}
    \caption{The covariance conditioning of the SVD meta-layer during the training process in the tasks of decorrelated BN (\emph{left}) and GCP (\emph{Right}). The decorrelated BN is based on ResNet-50 and CIFAR100, while ImageNet and ResNet-18 are used for the GCP. \vspace{-0.5cm}}
    \label{fig:cover_cond}
\end{figure}

This phenomenon was first observed in the GCP methods by~\cite{song2021approximate}, and we found that it generally extrapolates to other SVD-related tasks, such as decorrelated BN. Fig.~\ref{fig:cover_cond} depicts the covariance conditioning of these two tasks throughout the training. As can be seen, the integration of the SVD layer makes the generated covariance very ill-conditioned  (${\approx}1e12$ for decorrelated BN and ${\approx}1e16$ for GCP). By contrast, the conditioning of the approximate 
solver (Newton-Schulz iteration~\cite{higham2008functions}) is about $1e5$ for decorrelated BN and is around $1e15$ for GCP, while the standard BN only has a condition number of $1e3$.

%For decorrelated BN, the covariance 
%To avoid the huge time consumption of the SVD, some iterative methods are developed to approximate the solution~\cite{higham2008functions,song2022fast,song2022fast2}. 
%There are some approximate methods for the SVD
%A natural substitute of SVD is Newton-Schulz iteration (NS iteration)~\cite{higham2008functions} and other matrix approximation technique~\cite{song2022fast}.

% Explain the drawback and consequence of ill-conditioned covariance (FP+BP);
Ill-conditioned covariance matrices can harm the training of the network in both the forward pass (FP) and the backward pass (BP). For the FP, mainly the SVD solver is influenced in terms of stability and accuracy. Since the ill-conditioned covariance has many trivially-small eigenvalues, it is difficult for an SVD solver to accurately estimate them and large round-off errors are likely to be triggered, which might hurt the network performances. Moreover, the very imbalanced eigenvalue distribution can easily make the SVD solver fail to converge and cause the training failure~\cite{wang2021robust,song2021approximate}. For the BP, as pointed out in~\cite{lecun2012efficient,wiesler2011convergence,huang2018decorrelated}, the feature covariance is closely related to the Hessian matrix during the backpropagation. Since the error curvature is given by the eigenvalues of the Hessian matrix~\cite{sutskever2013importance}, for the ill-conditioned Hessian, the Gradient Descent (GD) step would bounce back and forth in high curvature directions (large eigenvalues) and make slow progress in low curvature directions (small eigenvalues). As a consequence, the ill-conditioned covariance could cause slow convergence and oscillations in the optimization landscape. 
The generalization abilities of a deep model are thus harmed.

% Our simplification: analyze the Pre-SVD layer in two consecutive training steps;
Due to the data-driven learning nature and the highly non-linear transform of deep neural networks, directly giving the analytical form of the covariance conditioning is intractable. Some simplifications have to be performed to ease the investigation. Since the covariance is generated and passed from the previous layer, the previous layer is likely to be the most relevant to the conditioning. Therefore, we naturally limit our focus to the Pre-SVD layer, \emph{i.e.,} the layer before the SVD layer. To further simplify the analysis, we study the Pre-SVD layer in two consecutive training steps, which can be considered as a mimic of the whole training process. Throughout the paper, we mainly investigate some meaningful manipulations on the weight, the gradient, and the learning rate of the Pre-SVD layer in two sequential training steps. \textit{Under our Pre-SVD layer simplifications, one promising direction to improve the conditioning is enforcing orthogonality on the weights.} Orthogonal weights have the norm-preserving property, which could improve the conditioning of the feature matrix. This technique has been widely studied in the literature of stable training and Lipschitz networks~\cite{mishkin2016all,wang2020orthogonal,singla2021skew}. We select some representative methods and validate their effectiveness in the task of decorrelated BN. Our experiment reveals that these orthogonal techniques can greatly improve the covariance conditioning, but could only bring marginal performance improvements and even slight degradation. \textit{This indicates that when the representation power of weight is limited, the improved conditioning does not necessarily lead to better performance. Orthogonalizing only the weight is thus insufficient to improve the generalization.}

%This also indicates that improved conditioning does not necessarily lead to better performance.  

% Our solution: nearest-orthogonal gradients and optimal learning rate;
% Moreover, the joint usage with the orthogonal weight brings the best results;
% The effectiveness is validated in two CV tasks.

%Under our experiments, we found that these techniques only brings marginal or even side effects on the performance. We conjecture that the orthogonality might put strong constraint on the weights and limit the representation power of the network. The performance therefore deteriorates. 
Instead of seeking orthogonality constraints on the weights, we propose our Nearest Orthogonal Gradient (NOG) and Optimal Learning Rate (OLR). These two techniques explore the orthogonality possibilities about the learning rate and the gradient. More specifically, our NOG modifies the gradient of the Pre-SVD layer into its nearest-orthogonal form and keeps the GD direction unchanged. On the other hand, the proposed OLR dynamically changes the learning rate of the Pre-SVD layer at each training step such that the updated weight is as close to an orthogonal matrix as possible. The experimental results demonstrate that the proposed two techniques not only significantly improve the covariance conditioning but also bring obvious improvements in the validation accuracy of both GCP and decorrelated BN. Moreover, when combined with the orthogonal weight treatments, the performance can have further improvements. 

The main contributions and findings are summarized below:
\begin{itemize}
    \item We systematically study the problem of how to improve the covariance conditioning of the SVD meta-layer. We propose our Pre-SVD layer simplification to investigate this problem from the perspective of orthogonal constraints. 
    \item We explore different techniques of orthogonal weights to improve the covariance conditioning. Our experiments reveal that these techniques could improve the conditioning but would harm the generalization abilities due to the limitation on the representation power of weight.
    \item We propose the nearest orthogonal gradient and optimal learning rate. The experiments on GCP and decorrelated BN demonstrate that these methods can attain better covariance conditioning and improved generalization. Their combinations with weight treatments can further boost the performance. 
\end{itemize}
\section{Related Work}

In this section, we introduce the related work in differentiable matrix decomposition and the orthogonality in neural networks which could be relevant in improving the covariance conditioning.

%, followed by a discussion on the orthogonality in neural network which could be relevant in improving the covariance conditioning.

\subsection{Differentiable Matrix Decomposition}

The differentiable matrix decomposition is widely used in neural networks as a spectral meta-layer. Ionescu~\emph{et al.}~\cite{ionescu2015matrix,ionescu2015training} first propose the theory of matrix back-propagation and laid a foundation for the follow-up research. In deep neural networks, the transformation of matrix square root and its inverse are often desired due to the appealing spectral property. Their applications cover a wide range of computer vision tasks~\cite{song2022fast,song2022fast2}. To avoid the huge time consumption of the SVD, some iterative methods are also developed to approximate the solution~\cite{higham2008functions,song2022fast,song2022fast2}.  In~\cite{huang2018decorrelated,chiu2019understanding,huang2019iterative,huang2020investigation,huang2021group,song2022fast}, the inverse square root is used in the ZCA whitening transform to whiten the feature map, which is also known as the decorrelated BN. The Global Covariance Pooling (GCP) models~\cite{li2017second,li2018towards,wang2020deep,xie2021so,song2021approximate,gao2021temporal,song2022eigenvalues} compute the matrix square root of the covariance as a spectral normalization, which achieves impressive performances on some recognition tasks, including large-scale visual classification~\cite{li2017second,song2021approximate,xie2021so,song2022fast}, fine-grained visual categorization~\cite{li2017second,li2018towards,song2022eigenvalues}, and video action recognition~\cite{gao2021temporal}. The Whitening and Coloring Transform (WCT), which uses both the matrix square root and inverse square root, is usually adopted in some image generation tasks such as neural style transfer~\cite{li2017universal,wang2020diversified}, image translation~\cite{ulyanov2017improved,cho2019image}, and domain adaptation~\cite{abramov2020keep,choi2021robustnet}. In the geometric vision problems, the differentiable SVD is usually applied to estimate the fundamental matrix and the camera pose~\cite{ranftl2018deep,dang2020eigendecomposition,campbell2020solving}. Besides the SVD-based factorization, differentiating Cholesky decomposition~\cite{murray2016differentiation} and some low-rank decomposition is used to approximate the attention mechanism~\cite{geng2020attention,xiong2021nystromformer,lu2021soft} or to learn the constrained representations~\cite{chan2015pcanet,yang2017admm}.  

\subsection{Orthogonality in Neural Network}

Orthogonal weights have the benefit of the norm-preserving property, \emph{i.e.,} the relation $||\mW\mA||_{\rm F}{=}||\mA||_{\rm F}$ holds for any orthogonal $\mW$. When it comes to deep neural networks, such a property can ensure that the signal stably propagates through deep networks without either exploding or vanishing gradients~\cite{bengio1994learning,glorot2010understanding}, which could speed up convergence and encourage robustness and generalization. In general, there are three ways to enforce orthogonality to a layer: orthogonal weight initialization~\cite{saxe2014exact,mishkin2016all,xiao2018dynamical}, orthogonal regularization~\cite{rodriguez2016regularizing,bansal2018can,qi2020deep,bansal2018can,wang2020orthogonal}, and explicit orthogonal weight via Carley transform or matrix exponential~\cite{maduranga2019complex,trockman2020orthogonalizing,singla2021skew}. Among these techniques, orthogonal regularization and orthogonal weight are most commonly used as they often bring some practical improvements in generalization. Since the covariance is closely related to the weight matrix of the Pre-SVD layer, enforcing the orthogonality constraint could help to improve the covariance conditioning of the SVD meta-layer. We will choose some representative methods and validate their impact in Sec.~\ref{sec:general_orthogonality}. 

Notice that the focus of existing literature is different from our work. The orthogonality constraints are often used to improve the Lipschitz constants of the neural network layers, which is expected to improve the visual quality in image generation~\cite{brock2018large,miyato2018spectral}, to allow for better adversarial robustness~\cite{tsuzuku2018lipschitz,singla2021skew}, and to improve generalization abilities~\cite{sedghi2018singular,wang2020orthogonal}. Our work is concerned with improving the covariance conditioning and generalization performance. Moreover, the orthogonality literature mainly investigates how to enforce orthogonality to weight matrices, whereas less attention is put on the gradient and learning rate. In Sec.~\ref{sec:NOG_OLR}, we will explore such possibilities and propose our solutions: nearest orthogonal gradient and optimal learning rate which is optimal in the sense that the updated weight is as close to an orthogonal matrix as possible. 

%and to build Lipschitz neural networks, which could leads to improved provable robustness against adversarial example. Our work does not involve adversarial robustness but is only concerned about better covariance conditioning and better standard generalization performance. Furthermore, the orthogonality literature mainly focuses on enforcing orthogonality to weight matrices, whereas less attention are put on the gradient and learning rate. In Sec.~\ref{sec:NOG_OLR}, we will study such possibilities and propose our solutions: nearest orthogonal gradient and orthogonality-optimal learning rate.

%So far, the orthogonal techniques are mainly designed for the weight matrices. 

%There are three ways to enforce orthogonality to the weight matrices. s

\section{Background: SVD Meta-Layer}

This section presents the background knowledge about the propagation rules of the SVD meta-layer.

%\subsection{}

\subsection{Forward Pass} 

Given the reshape feature $\mX{\in}\mathbb{R}^{d{\times}N}$ where $d$ denotes the feature dimensionality (\emph{i.e.,} the number of channels) and $N$ represents the number of features (\emph{i.e.,} the product of spatial dimensions of features), an SVD meta-layer first computes the sample covariance as:
\begin{equation}
    \mP = \mX \mJ \mX^{T}, \mJ =\frac{1}{N}(\mI-\frac{1}{N}\mathbf{1}\mathbf{1}^{T}) 
\end{equation}
where $\mJ$ represents the centering matrix, $\mathbf{I}$ denotes the identity matrix, and $\mathbf{1}$ is a column vector whose values are all ones, respectively. The covariance is always positive semi-definite (PSD) and does not have any negative eigenvalues. Afterward, the eigendecomposition is performed using the SVD:
\begin{equation}
    \mP=\mU\mathbf{\Lambda}\mU^{T},\ \mathbf{\Lambda}=\rm{diag}(\lambda_{1},\dots,\lambda_{d})
    \label{SVD}
\end{equation}
where $\mathbf{U}$ is the orthogonal eigenvector matrix, ${\rm diag}(\cdot)$ denotes transforming a vector to a diagonal matrix, and $\mathbf{\Lambda}$ is the diagonal matrix in which the eigenvalues are sorted in a non-increasing order \emph{i.e.}, $\lambda_i {\geq} \lambda_{i+1}$. Then depending on the application, the matrix square root or the inverse square root is calculated as:
\begin{equation}
\begin{gathered}
    \mathbf{Q}\triangleq\mathbf{P}^{\frac{1}{2}}=\mathbf{U}\mathbf{\Lambda}^{\frac{1}{2}} \mathbf{U}^{T}, \mathbf{\Lambda}^{\frac{1}{2}}={\rm diag}(\lambda_{1}^{\frac{1}{2}},\dots,\lambda_{d}^{\frac{1}{2}}) \\ 
    \mathbf{S}\triangleq\mathbf{P}^{-\frac{1}{2}}=\mathbf{U}\mathbf{\Lambda}^{-\frac{1}{2}} \mathbf{U}^{T}, \mathbf{\Lambda}^{-\frac{1}{2}}={\rm diag}(\lambda_{1}^{-\frac{1}{2}},\dots,\lambda_{d}^{-\frac{1}{2}})
\end{gathered}
\end{equation}
The matrix square root $\mQ$ is often used in GCP-related tasks~\cite{li2017second,xie2021so,song2021approximate}, while the application of decorrelated BN~\cite{huang2018decorrelated,siarohin2018whitening} widely applies the inverse square root $\mS$. In certain applications such as WCT, both $\mQ$ and $\mS$ are required. 

%for different feature transforms.

\subsection{Backward Pass} 

Let $\frac{\partial l}{\partial\mQ}$ and $\frac{\partial l}{\partial\mS}$ denote the partial derivative of the loss $l$ w.r.t to the matrix square root $\mQ$ and the inverse square root $\mS$, respectively. Then the gradient passed to the eigenvector is computed as:
\begin{equation}
    \frac{\partial l}{\partial \mathbf{U}}\Big|_{\mQ}=(\frac{\partial l}{\partial \mathbf{Q}} + (\frac{\partial l}{\partial \mathbf{Q}})^{T})\mathbf{U}\mathbf{\Lambda}^{\frac{1}{2}},\ 
    \frac{\partial l}{\partial \mathbf{U}}\Big|_{\mS}=(\frac{\partial l}{\partial \mathbf{S}} + (\frac{\partial l}{\partial \mathbf{S}})^{T})\mathbf{U}\mathbf{\Lambda}^{-\frac{1}{2}}
    \label{vec_de}
\end{equation}
Notice that the gradient equations for $\mQ$ and $\mS$ are different. For the eigenvalue, the gradient is calculated as:
\begin{equation}
\begin{gathered}
    \frac{\partial l}{\partial \mathbf{\Lambda}}\Big|_{\mQ}=\frac{1}{2}\rm{diag}(\lambda_{1}^{-\frac{1}{2}},\dots,\lambda_{d}^{-\frac{1}{2}})\mathbf{U}^{T} \frac{\partial \it{l}}{\partial \mathbf{Q}} \mathbf{U}, 
    \frac{\partial l}{\partial \mathbf{\Lambda}}\Big|_{\mS}=-\frac{1}{2}\rm{diag}(\lambda_{1}^{-\frac{3}{2}},\dots,\lambda_{d}^{-\frac{3}{2}})\mathbf{U}^{T} \frac{\partial \it{l}}{\partial \mathbf{S}} \mathbf{U}
\end{gathered}
\end{equation}
%where $\rm{diag}(\cdot)$ denotes the operation of transforming a vector into a diagonal matrix. 
Subsequently, the derivative of the SVD step can be calculated as:
\begin{equation}
    \frac{\partial l}{\partial \mathbf{P}}=\mathbf{U}( (\mathbf{K}^{T}\circ(\mathbf{U}^{T}\frac{\partial l}{\partial \mathbf{U}}))+ (\frac{\partial l}{\partial \mathbf{\Lambda}})_{\rm diag})\mathbf{U}^{T}
    \label{COV_de}
\end{equation}
where $\circ$ denotes the matrix Hadamard product, and the matrix $\mathbf{K}$ consists of entries $K_{ij}{=}{1}/{(\lambda_{i}{-}\lambda_{j})}$ if $i{\neq}j$ and $K_{ij}{=}0$ otherwise. This step is the same for both $\mQ$ and $\mS$. Finally, we have the gradient passed to the feature $\mX$ as:
\begin{equation}
    \frac{\partial l}{\partial \mathbf{X}}=(\frac{\partial l}{\partial \mathbf{P}}+(\frac{\partial l}{\partial \mathbf{P}})^{T})\mathbf{X}\mJ
    \label{X_de}
\end{equation}
With the above rules, the SVD function can be easily inserted into any neural networks and trained end-to-end as a meta-layer. 

%\subsection{Ill-conditioned Covariance.} 
%As discussed before, when the SVD meta-layer is integrated into a neural network, the covariance $\mP$ in eq.~\ref{SVD} is very likely to be ill-conditioned. 

\section{Pre-SVD Layer and Weight Treatments}

In this section, we first motivate our simplification of the Pre-SVD layer, and then validate the efficacy of some representative weight treatments.

\subsection{Pre-SVD Layer Simplification}
\label{sec:pre_svd}
The neural network consists of a sequential of non-linear layers where the learning of each layer is data-driven. Stacking these layers leads to a highly non-linear and complex transform, which makes directly analyzing the covariance conditioning intractable. To solve this issue, we have to perform some simplifications.  

Our simplifications involve limiting the analysis only to the layer previous to the SVD layer (which we dub as the Pre-SVD layer) in two consecutive training steps. The Pre-SVD layer directly determines the conditioning of the generated covariance, while the two successive training steps are a mimic of the whole training process. The idea is to simplify the complex transform by analyzing the sub-model (two layers) and the sub-training (two steps), which can be considered as an "abstract representation" of the deep model and its complete training.

Let $\mW$ denote the weight matrix of the Pre-SVD layer. Then for the input $\mX_{l}$ passed to the layer, we have:
\begin{equation}
    \mX_{l+1} = \mW\mX_{l} + \mathbf{b}
\end{equation}
where $\mX_{l+1}$ is the feature passed to the SVD layer, and $\mathbf{b}$ is the bias vector. %Notice that $\mW$ generally applies to any linear transformations including convolution. 
Since the bias $\mathbf{b}$ has a little influence here, we can sufficiently omit it for simplicity. The covariance in this step is computed as $\mW\mX_{l}\mX_{l}^{T}\mW^{T}$.
%The gradient passed to the weight is computed as:
%\begin{equation}
%    \frac{\partial l}{\partial \mathbf{W}} = \frac{\partial l}{\partial \mathbf{X}_{l+1}}\cdot \frac{\partial \mathbf{X}_{l+1}}{\partial \mathbf{W}}=\frac{\partial l}{\partial \mathbf{X}_{l+1}} \mathbf{X}_{l}^{T}
%\end{equation}
%Injecting this equation into eq.~\ref{X_de} leads to the re-formulation:
%\begin{equation}
%    \frac{\partial l}{\partial \mathbf{W}} = \Big(\frac{\partial l}{\partial \mathbf{P}}+(\frac{\partial l}{\partial \mathbf{P}})^{T}\Big)\mathbf{X}_{l}\mJ\mathbf{X}_{l-1}^{T}
%\end{equation}
After the BP, the weight matrix is updated as $\mathbf{W}{-}{\eta}\frac{\partial l}{\partial \mathbf{W}}$ where $\eta$ denotes the learning rate of the layer. Let $\mY_{l}$ denote the passed-in feature of the next training step. Then the covariance is calculated as:
\begin{equation}
\begin{aligned}
    \mC &= \Big( (\mathbf{W}-\eta\frac{\partial l}{\partial \mathbf{W}})\cdot\mY_{l} \Big)\Big( (\mathbf{W}-\eta\frac{\partial l}{\partial \mathbf{W}})\cdot\mY_{l} \Big)^{T}\\
    &=(\mathbf{W}-\eta\frac{\partial l}{\partial \mathbf{W}})\mY_{l}\mY_{l}^{T}(\mathbf{W}-\eta\frac{\partial l}{\partial \mathbf{W}})^{T}\\
    &=\mW\mY_{l}\mY_{l}^{T}\mW^{T} {-} \eta\frac{\partial l}{\partial \mathbf{W}}\mY_{l}\mY_{l}^{T}\mW^{T} {-} \eta\mW\mY_{l}\mY_{l}^{T}(\frac{\partial l}{\partial \mathbf{W}})^{T} {+} \eta^{2}\frac{\partial l}{\partial \mathbf{W}}\mY_{l}\mY_{l}^{T}(\frac{\partial l}{\partial \mathbf{W}})^{T}
    \label{eq:problem}
\end{aligned}
\end{equation}
where $\mC$ denotes the generated covariance of the second step. Now the problem becomes how to stop the new covariance $\mC$ from becoming worse-conditioned than $\mW\mX_{l}\mX_{l}^{T}\mW^{T}$. In eq.~\eqref{eq:problem}, three variables could influence the conditioning: the weight $\mW$, the gradient of the last step $\frac{\partial l}{\partial \mW}$, and the learning rate $\eta$ of this layer. Among them, the weight $\mW$ seems to be the most important as it contributes to three terms of eq.~\eqref{eq:problem}. Moreover, the first term $\mW\mY_{l}\mY_{l}^{T}\mW^{T}$ computed by $\mW$ is not attenuated by $\eta $ or $\eta^2$ like the other terms. Therefore, it is natural to first consider manipulating $\mW$ such that the conditioning of $\mC$ could be improved. 

\subsection{General Treatments on Weights}
\label{sec:general_orthogonality}

In the literature of enforcing orthogonality to the neural network, there are several techniques to improve the conditioning of the weight $\mW$. Now we introduce some representatives methods and validate their impacts. 

\subsubsection{Spectral Normalization (SN).} In~\cite{miyato2018spectral}, the authors propose a normalization method to stabilize the training of generative models~\cite{goodfellow2014generative} by dividing the weight matrix with its largest eigenvalue. The process is defined as:
\begin{equation}
    \mW / \sigma_{max}(\mW)
\end{equation}
Such a normalization can ensure that the spectral radius of $\mW$ is always $1$, \emph{i.e.,} $\sigma_{max}(\mW){=}1$. This could help to reduce the conditioning of the covariance since we have $\sigma_{max}(\mW\mY_{l}){=}\sigma_{max}(\mY_{l})$ after the spectral normalization.

%\subsubsection{Soft Orthogonality Regularization.} 
\subsubsection{Orthogonal Loss (OL).} Besides limiting the spectral radius of $\mW$, enforcing orthogonality constraint could also improve the covariance conditioning. As orthogonal matrices are norm-preserving (\emph{i.e.,} $||\mW\mY_{l}||_{\rm F}{=}||\mW||_{\rm F}$), lots of methods have been proposed to encourage orthogonality on weight matrices for more stable training and better signal-preserving property~\cite{pascanu2013difficulty,bansal2018can,wang2020orthogonal,trockman2020orthogonalizing,singla2021skew}. One common technique is to apply \emph{soft} orthogonality~\cite{wang2020orthogonal} by the following regularization:
\begin{equation}
    l=||\mW\mW^{T}-\mI||_{\rm F}
\end{equation}
This extra loss is added in the optimization objective to encourage more orthogonal weight matrices. However, since the constraint is achieved by regularization, the weight matrix is not exactly orthogonal at each training step.  

%\subsubsection{Hard Orthogonality Constraint.} 
\subsubsection{Orthogonal Weights (OW).} Instead of applying \emph{soft} orthogonality by regularization, some methods can explicitly enforce \emph{hard} orthogonality to the weight matrices~\cite{trockman2020orthogonalizing,singla2021skew}. The technique of~\cite{singla2021skew} is built on the mathematical property: for any skew-symmetric matrix, its matrix exponential is an orthogonal matrix.
\begin{equation}
    \exp(\mW-\mW^{T})\exp(\mW-\mW^{T})^{T}=\mI
\end{equation}
where the operation of $\mW{-}\mW^{T}$ is to make the matrix skew-symmetric, \emph{i.e.,} the relation $\mW{-}\mW^{T}{=}-(\mW{-}\mW^{T})^{T}$ always holds. Then $\exp(\mW{-}\mW^{T})$ is used as the weight. This technique explicitly constructs the weight as an orthogonal matrix. The orthogonal constraint is thus always satisfied during the training.

%Figure and table side by side

\begin{figure}\CenterFloatBoxes
\vspace{-0.4cm}
\begin{floatrow}
\ffigbox{%
  \includegraphics[width=0.99\linewidth]{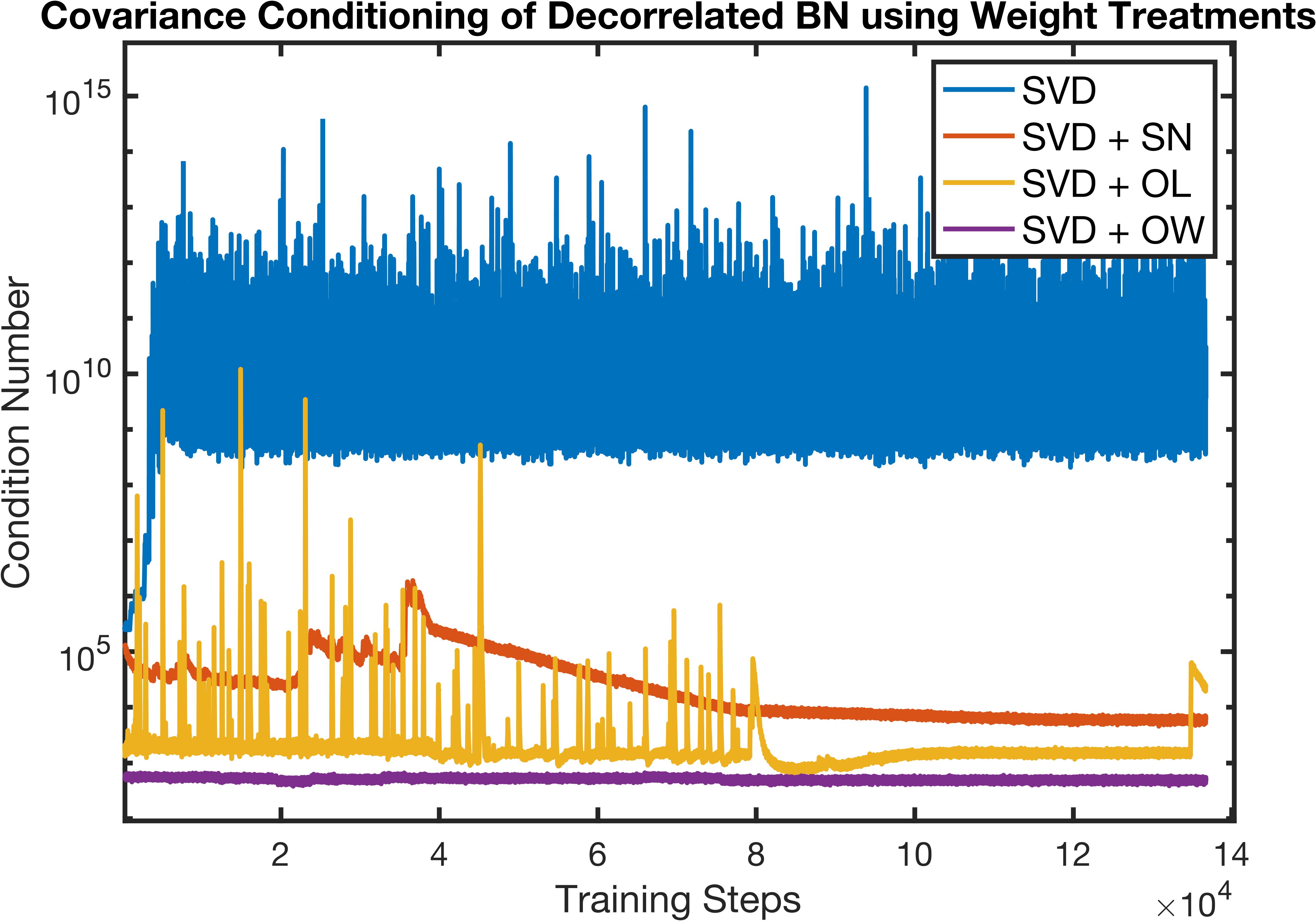}
}{%
  \caption{The covariance conditioning during the training process. All the weight treatments can improve the conditioning.}%
  \label{fig:ortho_weight}
}
\killfloatstyle\ttabbox[\Xhsize]{%
\resizebox{0.99\linewidth}{!}{
  \begin{tabular}{r|c|c} \toprule
  Methods & mean$\pm$std & min \\ \hline
  SVD & 19.99$\pm$0.16 &19.80 \\ \hline
  SVD + SN & 19.94$\pm$0.33 &19.60 \\
  SVD + OL & \textbf{19.73$\pm$0.28} & \textbf{19.54} \\
  SVD + OW & 20.06$\pm$0.17 &19.94 \\
  \hline\hline
    Newton-Schulz iteration &19.45$\pm$0.33&19.01\\
  \bottomrule
  \end{tabular}
}
}{%
  \caption{Performance of different weight treatments on ResNet-50 and CIFAR100 based on $10$ runs.}%
  \label{tab:ortho_weight}
}
\end{floatrow}
\vspace{-0.4cm}
\end{figure}

We apply the above three techniques in the experiment of decorrelated BN. Fig.~\ref{fig:ortho_weight} displays the covariance conditioning throughout the training, and Table~\ref{tab:ortho_weight} presents the corresponding validation errors. As can be seen, all of these techniques attain much better conditioning, but the performance improvements are not encouraging. The SN reduces the conditioning to around $10^{5}$, while the validation error marginally improves. The \emph{soft} orthogonality by the OL brings slight improvement on the performance despite some variations in the conditioning. The conditioning variations occur because the orthogonality constraint by regularization is not strictly enforced. Among the weight treatments, the \emph{hard} orthogonality by the OW achieves the best covariance conditioning, continuously maintaining the condition number around $10^{3}$ throughout the training. However, the OW slightly hurts the validation error. This implies that better covariance conditioning does not necessarily correspond to the improved performance, and orthogonalizing only the weight cannot improve the generalization. \textit{We conjecture that enforcing strict orthogonality only on the weight might limit its representation power.} Nonetheless, as will be discussed in Sec.~\ref{sec:NOG}, the side effect can be canceled when we simultaneously orthogonalize the gradient.

\section{Nearest Orthogonal Gradient \& Optimal Learning Rate} 
\label{sec:NOG_OLR}

In this section, we introduce our proposed two techniques on modifying the gradient and learning rate of the Pre-SVD layer. Their combinations with the weight treatments are also discussed.

\subsection{Nearest Orthogonal Gradient (NOG)}
\label{sec:NOG} 

As discussed in Sec.~\ref{sec:pre_svd}, the covariance conditioning is also influenced by the gradient $\frac{\partial l}{\partial \mW}$. However, existing literature mainly focuses on orthogonalizing the weights. To make the gradient also orthogonal, we propose to find the nearest-orthogonal gradient of the Pre-SVD layer. Different matrix nearness problems have been studied in~\cite{higham1988matrix}, and the nearest-orthogonal problem is defined as:
\begin{equation}
    \min_{\mR} ||\frac{\partial l}{\partial \mW}-\mR ||_{\rm F}\ subject\ to\ \mR\mR^{T}=\mI
\end{equation}
where $\mR$ is the seeking solution. To obtain such an orthogonal matrix, we can construct the error function as:
\begin{equation}
    e(\mR) = Tr\Big((\frac{\partial l}{\partial \mW}-\mR)^{T}(\frac{\partial l}{\partial \mW}-\mR)\Big) + Tr\Big(\mathbf{\Sigma} \mR^{T}\mR -\mI \Big) 
\end{equation}
where $Tr(\cdot)$ is the trace measure, and $\mathbf{\Sigma}$ denotes the symmetric matrix Lagrange multiplier. The closed-form solution is given by:
\begin{equation}
    \mR = \frac{\partial l}{\partial \mW} \Big(( \frac{\partial l}{\partial \mW})^{T} \frac{\partial l}{\partial \mW}\Big)^{-\frac{1}{2}}
\end{equation}
The detailed derivation is given in the supplementary material. If we have the SVD of the gradient ($\mU\mS\mV^{T}{=}\frac{\partial l}{\partial \mW}$), the solution can be further simplified as:
\begin{equation}
    \mR = \mU\mS\mV^{T} (\mV\mS^{-1}\mV^{T})=\mU\mV^{T} 
\end{equation}
As indicated above, the nearest orthogonal gradient is achieved by setting the singular value matrix to the identity matrix, \emph{i.e.,} setting $\mS$ to $\mI$. Notice that only the gradient of Pre-SVD layer is changed, while that of the other layers is not modified. Our proposed NOG can bring several practical benefits. 

\subsubsection{Orthogonal Constraint and Optimal Conditioning.} The orthogonal constraint is exactly enforced on the gradient as we have $(\mU\mV^{T})^{T}\mU\mV^{T}{=}\mI$. Since we explicitly set all the singular values to $1$, the optimal conditioning is also achieved, \emph{i.e.,} $\kappa(\frac{\partial l}{\partial \mW}){=}1$. This could help to improve the conditioning. 

%Notice that a general orthogonal matrix might not have this property because their eigenvalues could be $\pm1$.

\subsubsection{Keeping Gradient Descent Direction Unchanged.} In the high-dimensional optimization landscape, the many curvature directions (GD directions) are characterized by the eigenvectors of gradient ($\mU$ and $\mV$). Although our modification changes the gradient, the eigenvectors and the GD directions are untouched. In other words, our NOG only adjusts the step size in each GD direction. This indicates that the modified gradients will not harm the network performances. 

\subsubsection{Combination with Weight Treatments.} Our orthogonal gradient and the previous weight treatments are complementary. They can be jointly used to simultaneously orthogonalize the gradient and weight. In the following, we will validate their joint impact on the conditioning and performance.

%A natural approach is to combine our orthogonal gradients with other treatments on weights.

\begin{figure}\CenterFloatBoxes
\vspace{-0.4cm}
\begin{floatrow}
\ffigbox{%
  \includegraphics[width=0.99\linewidth]{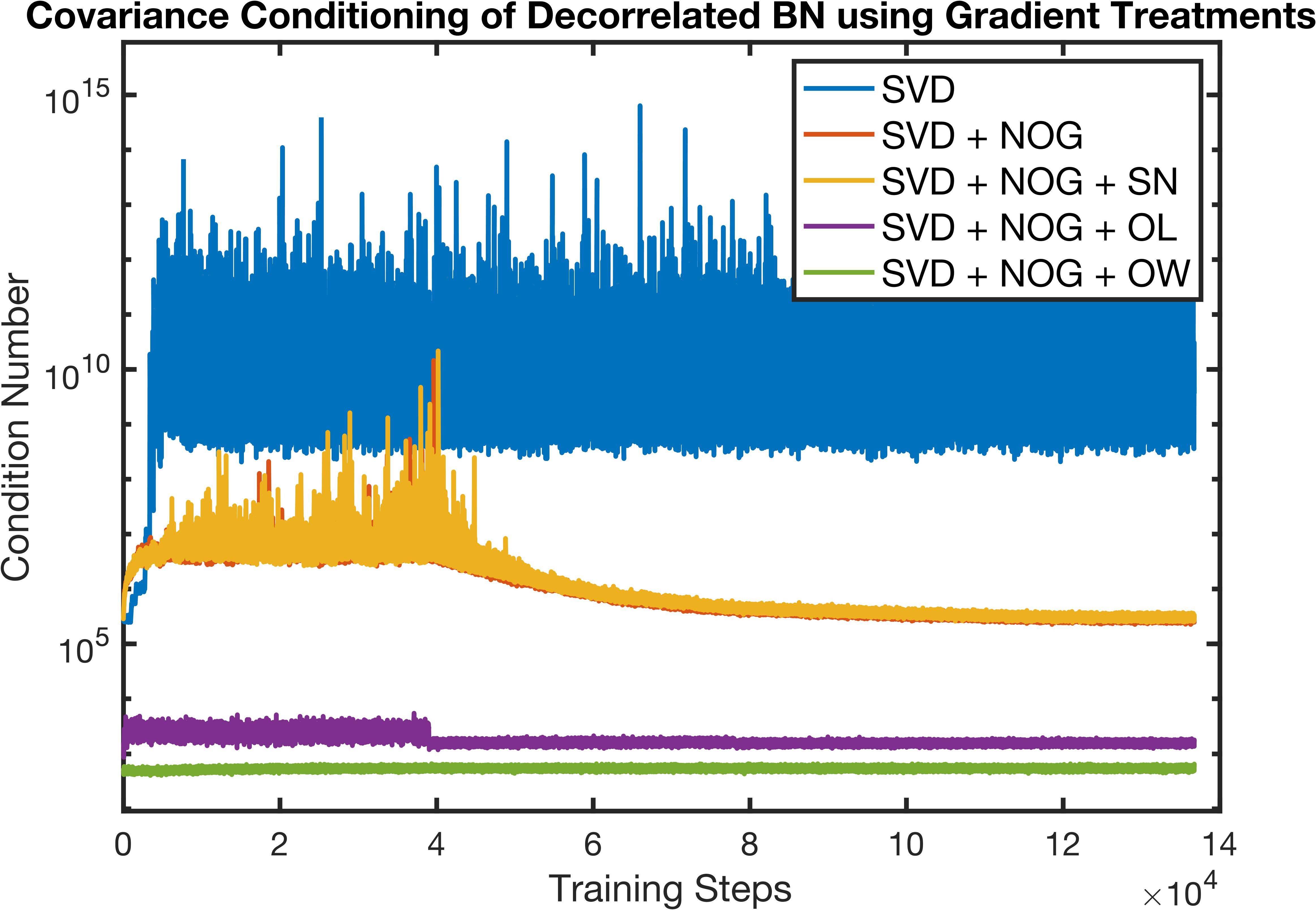}
}{%
  \caption{The covariance conditioning during the training process using orthogonal gradient and combined weight treatments.}%
  \label{fig:gradient}
}
\killfloatstyle\ttabbox[\Xhsize]{%
\resizebox{0.9\linewidth}{!}{
  \begin{tabular}{r|c|c} \toprule
  Methods & mean$\pm$std & min \\ \hline
  SVD & 19.99$\pm$0.16 &19.80 \\ 
  SVD + NOG & 19.43$\pm$0.24 &19.15\\\hline
  SVD + NOG + SN & 19.43$\pm$0.21 &19.20 \\
  SVD + NOG + OL & 20.14$\pm$0.39 &19.54 \\
  SVD + NOG + OW & \textbf{19.22$\pm$0.28}&\textbf{18.90} \\
  \hline\hline
    Newton-Schulz iteration &19.45$\pm$0.33&19.01\\
  \bottomrule
  \end{tabular}
}
}{%
  \caption{Performance of gradient and weight treatments on ResNet-50 and CIFAR100. Each result is based on $10$ runs.\vspace{-0.2cm}}%
  \label{tab:gradient}
}
\end{floatrow}
\end{figure}

Fig.~\ref{fig:gradient} and Table~\ref{tab:gradient} present the covariance conditioning of decorrelated BN and the corresponding validation errors, respectively. As we can observe, solely using the proposed NOG can largely improve the covariance conditioning, decreasing the condition number from $10^{12}$ to $10^6$. Though this improvement is not as significant as the orthogonal constraints (\emph{e.g.,} OL and OW), our NOG can benefit more the generalization abilities, leading to the improvement of validation error by $0.6\%$. Combining the SN with our NOG does not lead to obvious improvements in either the conditioning or validation errors, whereas the joint use of NOG and OL harms the network performances. This is because the orthogonality constraint by loss might not be enforced under the gradient manipulation. When our NOG is combined with the OW, the side effect of using only OW is eliminated and the performance is further boosted by $0.3\%$. This phenomenon demonstrates that when the gradient is orthogonal, applying the orthogonality constraint to the weight could also be beneficial to the generalization.

%outperforms only using NOG by $0.2\%$ and beats only using OW by $0.9\%$. 
%More specifically, jointing using NOG and OL outperforms using only OL by $0.5\%$, and combing NOG with OW significantly improves the performance by    

\subsection{Optimal Learning Rate (OLR)}
So far, we only consider orthogonalizing $\mW$ and $\frac{\partial l}{\partial \mW}$ separately, but 
how to jointly optimize $\mW{-}{\eta}\frac{\partial l}{\partial \mW}$ has not been studied yet. Actually, it is desired to choose an appropriate learning rate $\eta$ such that the updated weight is close to an orthogonal matrix. To this end, we need to achieve the following objective:
\begin{equation}
   \min_{\eta} ||(\mW-{\eta}\frac{\partial l}{\partial \mW})(\mW-{\eta}\frac{\partial l}{\partial \mW})^{T}-\mI||_{\rm F}
   %\min_{\eta} ||(\mW{-}{\eta}\frac{\partial l}{\partial \mW})||_{\rm F}
\end{equation}
This optimization problem can be more easily solved in the vector form. Let $\mathbf{w}$, $\mi$, and $\mathbf{l}$ denote the vectorized $\mW$, $\mI$, and $\frac{\partial l}{\partial \mW}$, respectively. Then we construct the error function as:
\begin{equation}
   e(\eta) = \Big((\mathbf{w}-\eta\mathbf{l})^{T}(\mathbf{w}-\eta\mathbf{l})-\mathbf{i}\Big)^{T}\Big((\mathbf{w}-\eta\mathbf{l})^{T}(\mathbf{w}-\eta\mathbf{l})-\mathbf{i}\Big)
\end{equation}
Expanding and differentiating the equation w.r.t. $\eta$ lead to:
\begin{equation}
\begin{gathered}
    \frac{d e(\eta)}{d \eta} \approx -4\mw\mw^{T}\ml^{T}\mw + 4\eta\mw\mw^{T}\ml^{T}\ml + 8\eta\ml^{T}\mw\ml^{T}\mw =0\\
   \eta^{\star} \approx \frac{\mw^{T}\mw\ml^{T}\mw}{\mw^{T}\mw\ml^{T}\ml+2\ml^{T}\mw\ml^{T}\mw}
   \label{optimal_lr}
\end{gathered}
\end{equation}
where some higher-order terms are neglected. The detailed derivation is given in the supplementary material. Though the proposed OLR yields the updated weight nearest to an orthogonal matrix theoretically, the value of $\eta^{\star}$ is unbounded for arbitrary $\mw$ and $\ml$. Directly using $\eta^{\star}$ might cause unstable training. To avoid this issue, we propose to use the OLR only when its value is smaller than the learning rate of other layers. Let $lr$ denote the learning rate of the other layers. The switch process can be defined as:
\begin{equation}
    \eta =\begin{cases}
    \eta^{\star} & if\ \eta^{\star}<lr\\
    lr & otherwise
    \end{cases}
\end{equation}

\subsubsection{Combination with Weight/Gradient Treatments.} When either the weight or the gradient is orthogonal, our OLR needs to be carefully used. When only $\mW$ is orthogonal, $\mw^{T}\mw$ is a small constant and it is very likely to have $\mw^{T}\mw{\ll}\ml^{T}\mw$. Consequently, we have $\mw^{T}\mw\ml^{T}\mw{\ll}\ml^{T}\mw\ml^{T}\mw$ and $\eta^{\star}$ will attenuate to zero. Similarly for orthogonal gradient, we have $\mw^{T}\mw\ml^{T}\mw{\ll}\ml^{T}\mw\ml^{T}\ml$ and this will cause $\eta^{\star}$ close to zero. Therefore, the proposed OLR cannot work when either the weight or gradient is orthogonal. Nonetheless, we note that if both $\mW$ and $\frac{\partial l}{\partial \mW}$ are orthogonal, our $\eta^{\star}$ is bounded. Specifically, we have:

\begin{prop}
 When both $\mW$ and $\frac{\partial l}{\partial \mW}$ are orthogonal, $\eta^{\star}$ is both upper and lower bounded. The upper bound is $\frac{N^2}{N^2 + 2}$ and the lower bound is $\frac{1}{N^{2}+2}$ where $N$ denotes the row dimension of $\mW$.
\end{prop}

We give the detailed proof in the supplementary material. Obviously, the upper bound of $\eta^{\star}$ is smaller than $1$. For the lower bound, since the row dimension of $N$ is often large (\emph{e.g.,} $64$), the lower bound of $\eta^{\star}$ can be according very small (\emph{e.g.,} $2e{-}4$). This indicates that our proposed OLR could also give a small learning rate even in the later stage of the training process.

%even in the regime of training network using a small learning rate. 

%We give the detailed proof in the Appendix. Only when $\mW$ and $\frac{\partial l}{\partial \mW}$ are both orthogonal, $\eta$ can have an upper-bounded estimation.  

In summary, the optimal learning rate is set such that the updated weight is optimal in the sense that it become as close to an orthogonal matrix as possible. In particular, it is suitable when both the gradient and weight are orthogonal.

%It is can be used when 

%By setting the learning rate this way, we can ensure that $\eta$ is optimal in the sense that the updated parameter is closest in the Frobenius norm to a orthogonal matrix. Notice that only the pre-SVD layer has varying learning rate at each training step, but learning rate of other layers remain unchanged. 

%Clearly this optimal learning rate can greatly change the optimization landscape and accelerate the convergence. However, it cannot be used in the later training stage when the learning rate of other layers is very small (\emph{e.g.,} $0.001$). Because the step size of other layers is very small, the varying learning rate of the pre-SVD layer would make the model oscillate around descent direction in the optimization landscape. Nonetheless, we find out that using the optimal and dynamic learning rate in the first few training epochs (\emph{e.g.,} 5) is sufficient to improve and maintain good covariance conditioning throughout the whole training process.

\begin{figure}\CenterFloatBoxes
\vspace{-0.4cm}
\begin{floatrow}
\ffigbox{%
  \includegraphics[width=0.9\linewidth]{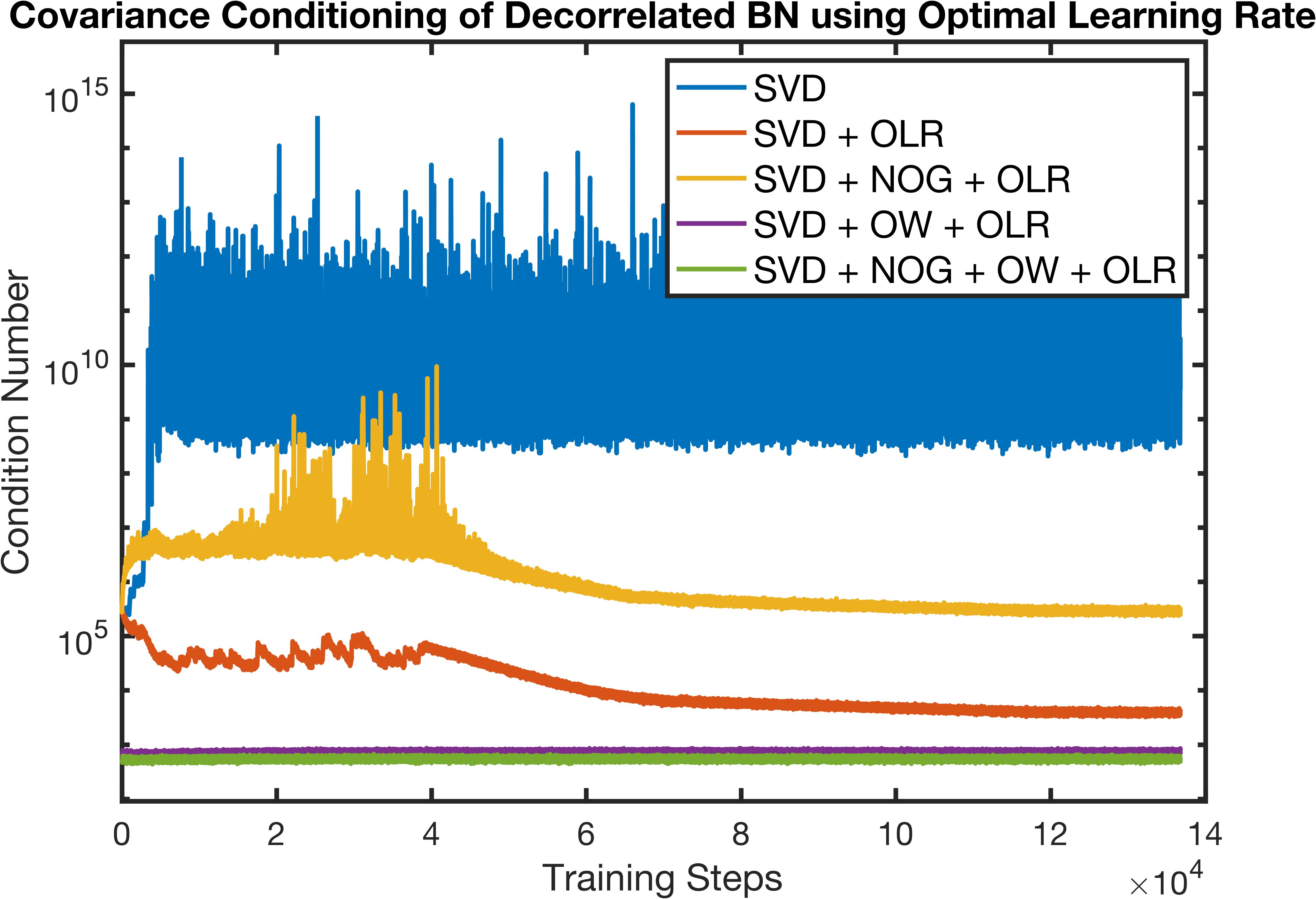}
}{%
  \caption{The covariance conditioning during the training process using optimal learning rate and hybrid treatments.\vspace{-0.2cm}}%
  \label{fig:olr}
}
\killfloatstyle\ttabbox[\Xhsize]{%
\resizebox{0.99\linewidth}{!}{
  \begin{tabular}{r|c|c} \toprule
  Methods & mean$\pm$std & min \\ \hline
  SVD & 19.99$\pm$0.16 &19.80 \\ 
  SVD + OLR & 19.50$\pm$0.39 &18.95 \\\hline
  SVD + NOG + OLR & 19.77$\pm$0.27  &19.36 \\
  SVD + OW + OLR  & 20.61$\pm$0.22  &20.43 \\
  SVD + NOG + OW +OLR & \textbf{19.05$\pm$0.31}&\textbf{18.77} \\
  \hline\hline
    Newton-Schulz iteration &19.45$\pm$0.33&19.01\\
  \bottomrule
  \end{tabular}
}
}{%
  \caption{Performance of optimal learning rate and hybrid treatments on ResNet-50 and CIFAR100 based on $10$ runs.\vspace{-0.2cm}}%
  \label{tab:olr}
}
\end{floatrow}
\vspace{-0.4cm}
\end{figure}

We give the covariance conditioning and the validation errors of our OLR in Fig.~\ref{fig:olr} and in Table~\ref{tab:olr}, respectively. Our proposed OLR significantly reduces the condition number to  $10^{4}$ and improves the validation error by $0.5\%$. When combined with either orthogonal weight or orthogonal gradient, there is a slight degradation on the validation errors. This meets our expectation as $\eta^{\star}$ would attenuate to zero in both cases. However, when both $\mW$ and $\frac{\partial l}{\partial \mW}$ are orthogonal, jointly using our OLR achieves the best performance, outperforming only OLR by $0.5\%$ and beating OW$+$NOG by $0.2\%$. This observation confirms that the proposed OLR works well for simultaneously orthogonal $\mW$ and $\frac{\partial l}{\partial \mW}$.

\section{Experiments}

We validate the proposed approaches in two applications: GCP and decorrelated BN. These two tasks are very representative because they have different usages of the SVD meta-layer. The GCP uses the matrix square root, while the decorrelated BN applies the inverse square root. In addition, the models of decorrelated BN often insert the SVD meta-layer at the beginning of the network, whereas the GCP models integrate the layer before the FC layer. 

\begin{table}[ht]
    \centering
    \resizebox{0.99\linewidth}{!}{
    \begin{tabular}{r|c|c|c|c}
    \toprule
         \multirow{2}*{Methods} & \multicolumn{2}{c|}{CIFAR10} & \multicolumn{2}{c}{CIFAR100} \\
         \cline{2-5}
         &mean$\pm$std & min &mean$\pm$std & min\\
         \hline
         SVD &4.35$\pm$0.09&4.17&19.99$\pm$0.16&19.80 \\
         %NS iteration &4.20$\pm$0.11 &4.11  &19.45$\pm$0.33&\textbf{\textcolor{cyan}{19.01}}\\
         \hline
         SVD + Spectral Norm (SN) &4.31$\pm$0.10 &4.15 &19.94$\pm$0.33 &19.60 \\
         SVD + Orthogonal Loss (OL) &4.28$\pm$0.07 &4.23 &19.73$\pm$0.28 &19.54\\
         SVD + Orthogonal Weight (OW) &4.42$\pm$0.09& 4.28&20.06$\pm$0.17&19.94\\
         \hline
         SVD +  Nearest Orthogonal Gradient (NOG) &\textbf{\textcolor{green}{4.15$\pm$0.06}}&\textbf{\textcolor{cyan}{4.04}} &\textbf{\textcolor{green}{19.43$\pm$0.24}} &\textbf{\textcolor{cyan}{19.15}} \\
         SVD + Optimal Learning Rate (OLR) &\textbf{\textcolor{cyan}{4.23$\pm$0.17}}&\textbf{\textcolor{blue}{3.98}} &\textbf{\textcolor{cyan}{19.50$\pm$0.39}}&\textbf{\textcolor{green}{18.95}} \\
         \hline
         SVD + NOG + OW &\textbf{\textcolor{blue}{4.09$\pm$0.07}}& \textbf{\textcolor{green}{4.01}} &\textbf{\textcolor{blue}{19.22$\pm$0.28}}&\textbf{\textcolor{blue}{18.90}} \\
         SVD + NOG + OW + OLR &\textbf{\textcolor{red}{3.93$\pm$0.09}}&\textbf{\textcolor{red}{3.85}} &\textbf{\textcolor{red}{19.05$\pm$0.31}}&\textbf{\textcolor{red}{18.77}} \\
    \hline\hline
    Newton-Schulz iteration &4.20$\pm$0.11 &4.11  &19.45$\pm$0.33&19.01\\
    \bottomrule
    \end{tabular}
    }
    \caption{Performance comparison of different decorrelated BN methods on CIFAR10/CIFAR100~\cite{krizhevsky2009learning} based on ResNet-50~\cite{he2016deep}. We report each result based on $10$ runs. The best four results are highlighted in \textbf{\textcolor{red}{red}}, \textbf{\textcolor{blue}{blue}}, \textbf{\textcolor{green}{green}}, and \textbf{\textcolor{cyan}{cyan}} respectively.}
    \label{tab:zca_res50}
    \vspace{-0.2cm}
\end{table}

\begin{table}[htbp]
    \centering
    \caption{Performance comparison of different GCP methods on ImageNet~\cite{deng2009imagenet} based on ResNet-18~\cite{he2016deep}. The failure times denote the total times of non-convergence of the SVD solver during one training process. The best four results are highlighted in \textbf{\textcolor{red}{red}}, \textbf{\textcolor{blue}{blue}}, \textbf{\textcolor{green}{green}}, and \textbf{\textcolor{cyan}{cyan}} respectively.}
    \resizebox{0.99\linewidth}{!}{
    \begin{tabular}{r|c|c|c}
    \toprule
        Method & Failure Times & Top-1 Acc. (\%) & Top-5 Acc. (\%) \\
    \hline
        SVD & 5 & 73.13 & 91.02 \\
    \hline
        SVD + Spectral Norm (SN) &2 &73.28 ($\uparrow$ 0.2) &91.11 ($\uparrow$ 0.1)\\
        SVD + Orthogonal Loss (OL) & 1& 71.75 ($\downarrow$ 1.4) &90.20 ($\downarrow$ 0.8)\\
        SVD + Orthogonal Weight (OW) &2 &73.07 ($\downarrow$ 0.1) &90.93 ($\downarrow$ 0.1)\\
    \hline
        SVD + Nearest Orthogonal Gradient (NOG) & 1 &\textbf{\textcolor{green}{73.51}} (\textbf{\textcolor{green}{$\uparrow$ 0.4}}) & \textbf{\textcolor{green}{91.35}} (\textbf{\textcolor{green}{$\uparrow$ 0.3}})\\
        SVD + Optimal Learning Rate (OLR) & 0 & \textbf{\textcolor{cyan}{73.39}} (\textbf{\textcolor{cyan}{$\uparrow$ 0.3}}) & \textbf{\textcolor{cyan}{91.26}} (\textbf{\textcolor{cyan}{$\uparrow$ 0.2}})\\
    \hline
        SVD + NOG + OW & 0  & \textbf{\textcolor{blue}{73.71}} (\textbf{\textcolor{blue}{$\uparrow$ 0.6}})& \textbf{\textcolor{blue}{91.43}} (\textbf{\textcolor{blue}{$\uparrow$ 0.4}})\\
        SVD + NOG + OW + OLR & 0  & \textbf{\textcolor{red}{73.82}} (\textbf{\textcolor{red}{$\uparrow$ 0.7}})& \textbf{\textcolor{red}{91.57}} (\textbf{\textcolor{red}{$\uparrow$ 0.6}})\\
    \hline\hline
    Newton-Schulz iteration & 0 & 73.36 ($\uparrow$ 0.2) & 90.96 ($\downarrow$ 0.1)\\
    \bottomrule
    \end{tabular}
    }
    \label{tab:gcp_res18}
    \vspace{-0.5cm}
\end{table}

\subsection{Decorrelated Batch Normalization}

Table~\ref{tab:zca_res50} compares the performance of each method on CIFAR10/CIFAR100~\cite{krizhevsky2009learning} based on ResNet-50~\cite{he2016deep}. Both of our NOG and OLR achieve better performance than other weight treatments and the SVD. Moreover, when hybrid treatments are adopted, we can observe step-wise steady improvements on the validation errors. Among these techniques, the joint usage of OLR with NOG and OW achieves the best performances across metrics and datasets, outperforming the SVD baseline by $0.4\%$ on CIFAR10 and by $0.9\%$ on CIFAR100. This demonstrates that these treatments are complementary and can benefit each other.

\subsection{Global Covariance Pooling}

Table~\ref{tab:gcp_res18} presents the total failure times of the SVD solver in one training process and the validation accuracy on ImageNet~\cite{deng2009imagenet} based on ResNet-18~\cite{he2016deep}. The results are very coherent with our experiment of decorrelated BN. Among the weight treatments, the OL and OW hurt the performance, while the SN improves that of SVD by $0.2\%$. Our proposed NOG and OLR outperform the weight treatments and improve the SVD baseline by $0.4\%$ and by $0.3\%$, respectively. Moreover, the combinations with the orthogonal weight further boost the performance. Specifically, combining NOG and OW surpasses the SVD by $0.6\%$. The joint use of OW with NOG and OLR achieves the best performance among all the methods and beats the SVD by $0.7\%$.

\begin{figure}[htbp]
  \vspace{-0.5cm}
    \centering
    \includegraphics[width=0.7\linewidth]{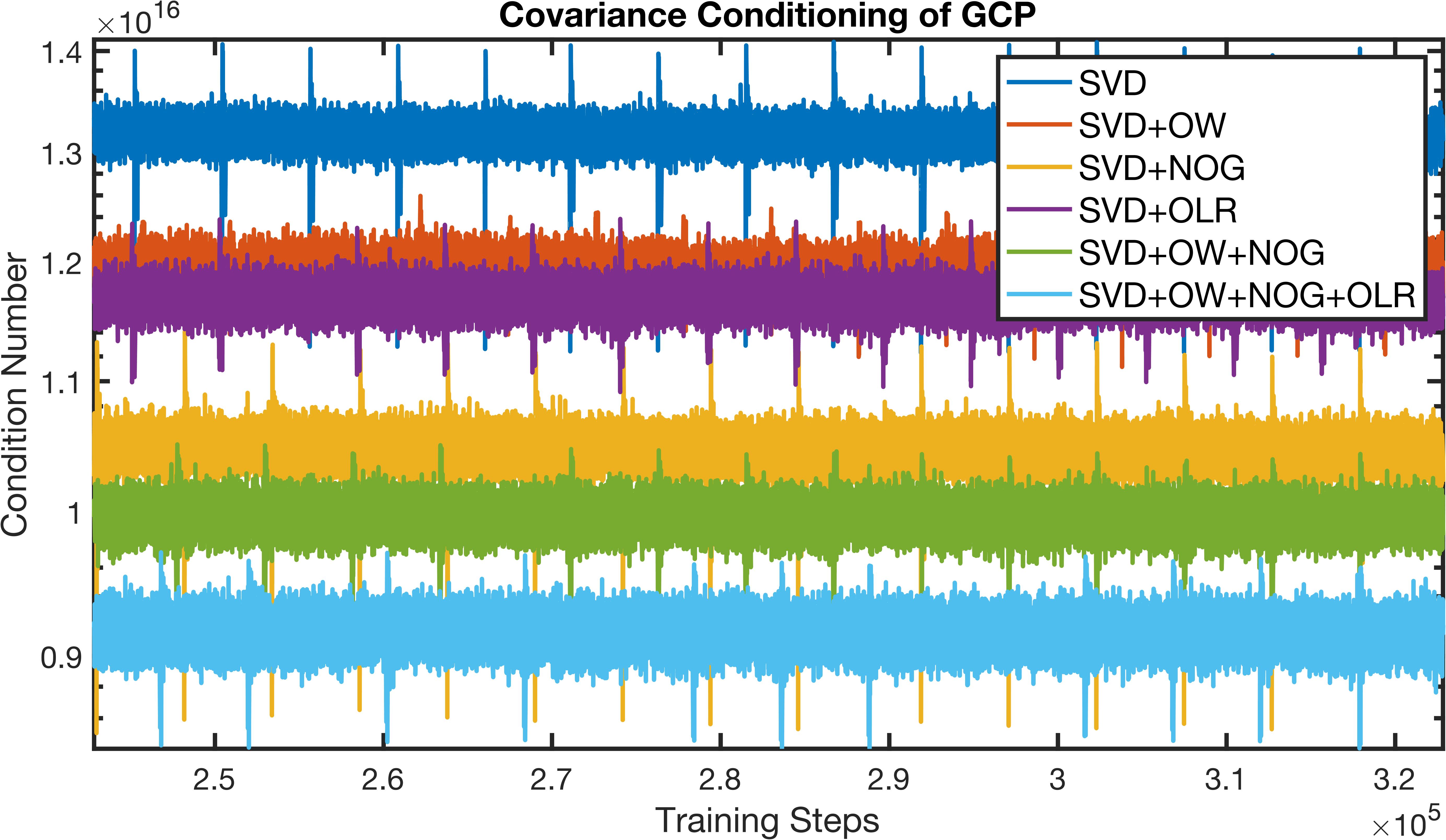}
    \caption{The covariance conditioning of GCP methods in the later stage of the training. The periodic spikes are caused by the evaluation on the validation set after every epoch.}
    \label{fig:gcp_cond}
    \vspace{-0.3cm}
\end{figure}

Fig.~\ref{fig:gcp_cond} depicts the covariance conditioning in the later training stage. Our OLR and the OW both reduce the condition number by around $1e15$, whereas the proposed NOG improves the condition number by $2e15$. When hybrid treatments are used, combining NOG and OW attains better conditioning than the separate usages. Furthermore, simultaneously using all the techniques leads to the best conditioning and improves the condition number by $5e15$. 

The covariance conditioning of GCP tasks is not improved as much as that of decorrelated BN. This might stem from the unique architecture of GCP models: the covariance is directly used as the final representation and fed to the FC layer. We conjecture that this setup might cause the covariance to have
a high condition number. The approximate solver (Newton-Schulz iteration) does not have well-conditioned matrices either (${\approx}1e15$), which partly supports our conjecture.

%which means that the improvement space is limited.  
%\subsubsection{Why Covariance Conditioning of GCP is only marginally improved.} Different from the experiment of decorrelated BN, the covariance conditioning is marginally improved on GCP tasks. This might stem from the fact the covariance of GCP is directly used as the final representation and fed to the FC layer. We conjecture that the representation power might be related to the conditioning: the feature with larger conditioning has more expressive power. Also, the approximate solver (Newton-Schulz iteration) does not have well-conditioned matrices either, which means the improvement space is limited.  

%\subsection{Neural Style Transfer}

%\subsection{Ablation Study}

%\subsection{Additional Computational Complexity}
\section{Conclusion and Future Work}

In this paper, we explore different approaches to improve the covariance conditioning of the SVD meta-layer. Existing treatments on orthogonal weight are first studied. Our experiments reveal that these techniques could improve the conditioning but might hurt the performance due to the limitation on the representation power. To avoid the side effect of orthogonal weight, we propose the nearest orthogonal gradient and the optimal learning rate, both of which could simultaneously attain better covariance conditioning and improved generalization abilities. Moreover, their combinations with orthogonal weight further boost the performance. The proposed orthogonal approaches have a direct beneficial influence on a wide variety of computer vision applications and might inspire other orthogonal techniques. In future work, we would like to study the problem of ill-conditioned covariance from other perspectives and extend our proposed techniques to other SVD-related methods. 

%It would also be interesting to investigate whether our treatments

%In the end of the paper, we do not want to pose specific suggestions on using what techniques to improve the covariance conditioning. 
%\subsection{Does better conditioning correspond to better generalization abilities?}

\clearpage
\bibliographystyle{splncs04}
\bibliography{egbib}

%\clearpage
\appendix
\section{Mathematical Derivation and Proof}

\subsection{Derivation of Nearest Orthogonal Gradient}

The problem of finding the nearest orthogonal gradient can be defined as:
\begin{equation}
    \min_{\mR} ||\frac{\partial l}{\partial \mW}-\mR ||_{\rm F}\ subject\ to\ \mR\mR^{T}=\mI
\end{equation}
To solve this constrained optimization problem, We can construct the following error function:
\begin{equation}
    e(\mR) = Tr\Big((\frac{\partial l}{\partial \mW}-\mR)^{T}(\frac{\partial l}{\partial \mW}-\mR)\Big) + Tr\Big(\mathbf{\Sigma} \mR^{T}\mR -\mI \Big) 
\end{equation}
where $Tr(\cdot)$ is the trace measure, and $\mathbf{\Sigma}$ denotes the symmetric matrix Lagrange multiplier. Setting the derivative to zero leads to:
\begin{equation}
\begin{gathered}
   \frac{d e(\mR)}{d \mR} = -2 (\frac{\partial l}{\partial \mW}-\mR) + 2\mR\mathbf{\Sigma} = 0 \\
   \frac{\partial l}{\partial \mW} = \mR(\mI + \mathbf{\Sigma} ),\  \mR = \frac{\partial l}{\partial \mW}(\mI + \mathbf{\Sigma})^{-1}
   \label{inter_result}
\end{gathered}
\end{equation}
The term $(\mI + \mathbf{\Sigma})$ can be represented using $\frac{\partial l}{\partial \mW}$. Consider the covariance of $\frac{\partial l}{\partial \mW}$:
\begin{equation}
\begin{gathered}
   (\frac{\partial l}{\partial \mW})^{T}\frac{\partial l}{\partial \mW} = (\mI + \mathbf{\Sigma} )^{T}\mR^{T}\mR(\mI + \mathbf{\Sigma} ) = (\mI + \mathbf{\Sigma} )^{T}(\mI + \mathbf{\Sigma} ) \\
   (\mI + \mathbf{\Sigma} ) = \Big((\frac{\partial l}{\partial \mW})^{T}\frac{\partial l}{\partial \mW}\Big)^{\frac{1}{2}}
\end{gathered}
\end{equation}
Substituting the term $(\mI + \mathbf{\Sigma})$ in eq.~\eqref{inter_result} with the above equation leads to the closed-form solution of the nearest orthogonal gradient:
\begin{equation}
    \mR = \frac{\partial l}{\partial \mW} \Big(( \frac{\partial l}{\partial \mW})^{T} \frac{\partial l}{\partial \mW}\Big)^{-\frac{1}{2}}
\end{equation}

\subsection{Derivation of Optimal Learning Rate}
To jointly optimize the updated weight $\mW{-}{\eta}\frac{\partial l}{\partial \mW}$, we need to achieve the following objective:
\begin{equation}
   \min_{\eta} ||(\mW{-}{\eta}\frac{\partial l}{\partial \mW})(\mW{-}{\eta}\frac{\partial l}{\partial \mW})^{T}-\mI||_{\rm F}
   %\min_{\eta} ||(\mW{-}{\eta}\frac{\partial l}{\partial \mW})||_{\rm F}
\end{equation}
This optimization problem can be more easily solved in the form of vector. Let $\mathbf{w}$, $\mi$, and $\mathbf{l}$ denote the vectorized $\mW$, $\mI$, and $\frac{\partial l}{\partial \mW}$, respectively. Then we construct the error function as:
\begin{equation}
   e(\eta) =\Big( (\mathbf{w}-\eta\mathbf{l})^{T}(\mathbf{w}-\eta\mathbf{l})-\mi\Big)^{T}\Big( (\mathbf{w}-\eta\mathbf{l})^{T}(\mathbf{w}-\eta\mathbf{l})-\mi\Big)
\end{equation}
Expanding the equation leads to:
\begin{equation}
     e(\eta)=(\mathbf{w}^{T}\mathbf{w}-2\eta\mathbf{l}^{T}\mathbf{w}+\eta^{2}\mathbf{l}^{T}\mathbf{l}-\mathbf{i})^{T}(\mathbf{w}^{T}\mathbf{w}-2\eta\mathbf{l}^{T}\mathbf{w}+\eta^{2}\mathbf{l}^{T}\mathbf{l}-\mathbf{i})
\end{equation}
Differentiating $e(\eta)$ w.r.t. $\eta$ yields:
\begin{equation}
\begin{gathered}
   \frac{d e(\eta)}{d \eta} = -4\mw\mw^{T}\ml^{T}\mw+ 4\eta\mw\mw^{T}\ml^{T}\ml+ 8\eta\ml^{T}\mw\ml^{T}\mw-12\eta^{2}\ml^{T}\mw\ml^{T}\ml+4\ml\mw^{T}\mi\\
   +4\eta^{3}\ml\ml^{T} - 4\eta\mi\ml\ml^{T}
\end{gathered}
\end{equation}
Since $\eta$ is typically very small, the higher-order terms (\emph{e.g.,} $\eta^{2}$ and $\eta^{3}$) are sufficiently small such that they can be neglected. After omitting these terms, the derivative becomes:
\begin{equation}
    \frac{d e(\eta)}{d \eta} \approx -4\mw\mw^{T}\ml^{T}\mw + 4\eta\mw\mw^{T}\ml^{T}\ml + 8\eta\ml^{T}\mw\ml^{T}\mw +4\ml\mw^{T}\mi - 4\eta\mi\ml\ml^{T}\\
\end{equation}
Setting the derivative to zero leads to the optimal learning rate:
\begin{equation}
    \eta^{\star} \approx \frac{\mw^{T}\mw\ml^{T}\mw-\ml^{T}\mw\mi}{\mw^{T}\mw\ml^{T}\ml+2\ml^{T}\mw\ml^{T}\mw - \ml^{T}\ml\mi}
\end{equation}
Notice that $\mi$ is the vectorization of the identify matrix $\mI$, which means that $\mi$ is very sparse ($\emph{i.e.,}$ lots of zeros) and the impact can be neglected. The optimal learning rate can be further simplified as:
\begin{equation}
    \eta^{\star} \approx \frac{\mw^{T}\mw\ml^{T}\mw}{\mw^{T}\mw\ml^{T}\ml+2\ml^{T}\mw\ml^{T}\mw}
\end{equation}

%Because $\mi$ is very space, the impact can be also neglected.

%We can construct the measure as:
%\begin{equation}
%    e(\eta) = Tr( (\mW{-}{\eta}\frac{\partial l}{\partial \mW})(\mW{-}{\eta}\frac{\partial l}{\partial \mW})^{T} -\mI)
%\end{equation}

\subsection{Proof of the learning rate bounds}

\begin{duplicate}
 When both $\mW$ and $\frac{\partial l}{\partial \mW}$ are orthogonal, $\eta^{\star}$ is both upper and lower bounded. The upper bound is $\frac{N^2}{N^2 + 2}$ and the lower bound is $\frac{1}{N^{2}+2}$ where $N$ denotes the row dimension of $\mW$.
\end{duplicate}
\begin{proof}

  Since the vector product is equivalent to the matrix Frobenius inner product, we have the relation: 
  \begin{equation}
       \ml^{T}\mw = \langle\frac{\partial l}{\partial \mW},\mW\rangle_{\rm F}
  \end{equation}
  For a given matrix pair $\mA$ and $\mB$, the Frobenius product $\langle\cdot\rangle_{\rm F}$ is defined as:
  \begin{equation}
      \langle\mA,\mB\rangle_{\rm F}=\sum A_{i,j}B_{i,j}\leq \sigma_{1}(\mA)\sigma_{1}(\mB)+\dots+\sigma_{N}(\mA)\sigma_{N}(\mB)
  \end{equation}
  where $\sigma(\cdot)_{i}$ represents the $i$-th largest eigenvalue, $N$ denotes the matrix size, and the inequality is given by Von Neumann’s trace inequality~\cite{mirsky1975trace,grigorieff1991note}. The equality takes only when $\mA$ and $\mB$ have the same eigenvector. When both $\mW$ and $\frac{\partial l}{\partial \mW}$ are orthogonal, \emph{i.e.,} their eigenvalues are all $1$, we have the following relation:
  \begin{equation}
      \langle\frac{\partial l}{\partial \mW},\frac{\partial l}{\partial \mW}\rangle_{\rm F}=N,\ 
      \langle\frac{\partial l}{\partial \mW},\mW\rangle_{\rm F}\leq N
  \end{equation}
  This directly leads to:
  \begin{equation}
      \langle\frac{\partial l}{\partial \mW},\mW\rangle_{\rm F}\leq\langle\frac{\partial l}{\partial \mW},\frac{\partial l}{\partial \mW}\rangle_{\rm F},\ \ml^{T}\mw \leq \ml^{T}\ml 
  \end{equation}
  Exploiting this inequality, the optimal learning rate has the relation:
  \begin{equation}
      \eta^{\star} \approx \frac{\mw^{T}\mw\ml^{T}\mw}{\mw^{T}\mw\ml^{T}\ml+2\ml^{T}\mw\ml^{T}\mw}
      \leq \frac{\mw^{T}\mw\ml^{T}\ml}{\mw^{T}\mw\ml^{T}\ml+2\ml^{T}\mw\ml^{T}\mw}
      \label{eq:upper_bound_1}
  \end{equation}
  For $\ml^{T}\mw$, we have the inequality as:
  \begin{equation}
      \ml^{T}\mw = \langle\frac{\partial l}{\partial \mW},\mW\rangle_{\rm F}=\sum_{i,j} \frac{\partial l}{\partial \mW}_{i,j}\mW_{i,j} \geq \sigma_{min}(\frac{\partial l}{\partial \mW})\sigma_{min}(\mW)=1
      \label{eq:inequality_lw}
  \end{equation}
  Then we have the upper bounded of $\eta^{\star}$ as:
  \begin{equation}
     \eta^{\star} \leq \frac{\mw^{T}\mw\ml^{T}\ml}{\mw^{T}\mw\ml^{T}\ml+2\ml^{T}\mw\ml^{T}\mw} = \frac{N^2}{N^2+2\ml^{T}\mw\ml^{T}\mw} < \frac{N^2}{N^2 + 2}
  \end{equation}
  For the lower bound, since we also have $\ml^{T}\mw{\leq}\mw^{T}\mw$, $\eta^{\star}$ can be re-written as:
  \begin{equation}
  \begin{gathered}
      \eta^{\star} \approx \frac{\mw^{T}\mw\ml^{T}\mw}{\mw^{T}\mw\ml^{T}\ml+2\ml^{T}\mw\ml^{T}\mw} \geq \frac{\ml^{T}\mw\ml^{T}\mw}{\mw^{T}\mw\ml^{T}\ml+2\ml^{T}\mw\ml^{T}\mw} = \frac{1}{\frac{\mw^{T}\mw\ml^{T}\ml}{\ml^{T}\mw\ml^{T}\mw}+2} 
      =\frac{1}{\frac{N^{2}}{\ml^{T}\mw\ml^{T}\mw}+2}
  \end{gathered}
  \label{eq:lower_bound_1}
  \end{equation}
  Injecting eq.~\eqref{eq:inequality_lw} into eq.~\eqref{eq:lower_bound_1} leads to the further simplification:
  \begin{equation}
     \eta^{\star} \approx \frac{1}{\frac{N^{2}}{\ml^{T}\mw\ml^{T}\mw}+2} \geq \frac{1}{N^{2}+2}
  \end{equation}
  As indicated above, the optimal learning rate $\eta^{\star}$ has a lower bound of $\frac{1}{N^{2}+2}$.
\end{proof}

%\subsection{Optimal Learning Rate + Orthogonal Weights}

%\subsection{Optimal Learning Rate + Orthogonal Gradients}

\section{Detailed Experimental Settings}

In this section, we introduce the implementation details and experimental settings of the two experiments.

\subsection{Decorrelated Batch Normalization}

\begin{figure}[htbp]
    \centering
    \includegraphics[width=0.4\linewidth]{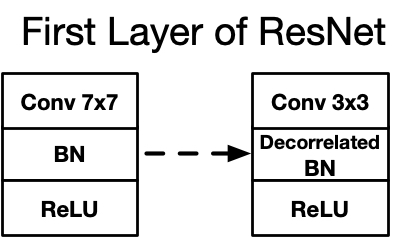}
    \caption{The scheme of the modified ResNet for decorrelated BN. We reduce the kernel size of the first convolution layer from $7{\times}7$ to $3{\times}3$. The BN after this layer is replaced with our decorrelated BN layer.}
    \label{fig:zca_arch}
\end{figure}

We use ResNet-50~\cite{he2016deep} as the backbone for the experiment on CIFAR10 and CIFAR100~\cite{krizhevsky2009learning}. The kernel size of the first convolution layer of ResNet is $7{\times}7$, which might not suit the low resolution of these two datasets (the images are only of size $32{\times}32$). To avoid this issue, we reduce the kernel size of the first convolution layer to $3{\times}3$. The stride is also decreased from $2$ to $1$. The BN layer after this layer is replace with our decorrelated BN layer (see Fig.~\ref{fig:zca_arch}). Let $\mX{\in}\mathbb{R}^{C{\times}BHW}$ denotes the reshaped feature. The whitening transform is performed as:
\begin{equation}
    \mX_{whitened} = (\mX\mX^{T})^{-\frac{1}{2}} \mX
\end{equation}
%where $\epsilon$ is a small constant to make sure that the covariance always have positive eigenvalues. We set $\epsilon$ to $0.01$ in our experiments. 
Compared with the vanilla BN that only standardizes the data, the decorrelated BN can further eliminate the data correlation between each dimension.

The training lasts $350$ epochs and the learning rate is initialized with $0.1$. The SGD optimizer is used with momentum $0.9$ and weight decay $5e{-}4$. We decrease the learning rate by $10$ every $100$ epochs. The batch size is set to $128$. We use the technique proposed in~\cite{song2021approximate} to compute the stable SVD gradient. The Pre-SVD layer in this experiment is the $3{\times}3$ convolution layer.

\subsection{Global Covariance Pooling}

\begin{figure}[htbp]
    \centering
    \includegraphics[width=0.8\linewidth]{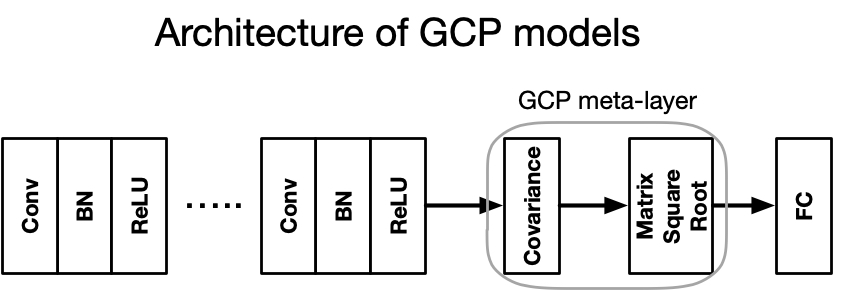}
    \caption{The architecture of a GCP model~\cite{li2017second,song2021approximate}. After all the convolution layers, the covariance square root of the feature is computed and used as the final representation.}
    \label{fig:gcp_arch}
\end{figure}

We use ResNet-18~\cite{he2016deep} for the GCP experiment and train it from scratch on ImageNet~\cite{deng2009imagenet}. Fig.~\ref{fig:gcp_arch} displays the overview of a GCP model. For the ResNet backbone, the last Global Average Pooling (GAP) layer is replaced with our GCP layer. Consider the final batched convolutional feature $\mX{\in}\mathbb{R}^{B{\times}C{\times}HW}$. We compute the matrix square root of its covariance as:
\begin{equation}
    \mQ = (\mX\mX^{T})^{\frac{1}{2}}
\end{equation}
where $\mQ{\in}\mathbb{R}^{B{\times}C{\times}C}$ is used as the final representation and directly passed to the fully-connected (FC) layer.   

The training process lasts $60$ epochs and the learning rate is initialize with $0.1$. We decrease the learning rate by $10$ at epoch $30$ and epoch $45$. The SGD optimizer is used with momentum $0.9$ and weight decay $1e{-}4$. The model weights are randomly initialized and the batch size is set to $256$. The images are first resized to $256{\times}256$ and then randomly cropped to $224{\times}224$ before being passed to the model. The data augmentation of randomly horizontal flip is used. We use the technique proposed in~\cite{song2021approximate} to compute the stable SVD gradient. The Pre-SVD layer denotes the convolution transform of the previous layer.

\section{Computational Cost}

\begin{table}[htbp]
    \centering
    \caption{Time consumption of each forward pass (FP) and backward pass (BP) measured on a RTX A6000 GPU. The evaluation is based on ResNet-50 and CIFAR100.} %The covariance size is $1{\times}64{\times}64$ for DBN and is $256{\times}196{\times}196$ for GCP.}
    \vspace{-0.1cm}
    %\resizebox{0.99\linewidth}{!}{
    \begin{tabular}{c|cc}
    \toprule
         Methods & FP (ms) & BP (ms) \\
    \midrule
         SVD & 44 & 95 \\
         SVD + NOG & 44 & 97 (+2)\\
         SVD + OLR & 44 & 96 (+1) \\
         SVD + OW & 48 (+4) & 102 (+7) \\
         SVD + OW + NOG + OLR & 49 (+5) &106 (+11) \\
         Newton-Schulz Iteration & 43 & 93\\
         \midrule
         Vanilla ResNet-50 & 42  & 90 \\
    \bottomrule
    \end{tabular}
    %}
    \label{tab:time}
\end{table}

Table~\ref{tab:time} compares the time consumption of a single training step for the experiment of decorrelated BN. Our NOG and OLR bring negligible computational costs to the BP ($2\%$ and $1\%$), while the FP is not influenced. Even when all techniques are applied, the overall time costs are marginally increased by $10\%$. Notice that NOG and OLR have no impact on the inference speed. 

\end{document}